\PassOptionsToPackage{table}{xcolor}
\documentclass[sigconf]{acmart}

\AtBeginDocument{%
  }

\setcopyright{acmlicensed}
\copyrightyear{2026}
\acmYear{2026}
\acmDOI{XXXXXXX.XXXXXXX}
\acmConference[MM '26]{Proceedings of the 34th ACM International Conference on Multimedia}{November 10-14, 2026}{Rio de Janeiro, Brazil}
\acmISBN{978-1-4503-XXXX-X/2018/06}

\usepackage{cleveref}
\usepackage{amsmath}

\usepackage{amssymb}
\usepackage{mathtools}
\usepackage{amsthm}
\usepackage{siunitx}
\usepackage{multirow}
\usepackage{makecell}
\usepackage{colortbl}
\definecolor{whiteshade}{gray}{1}
\definecolor{rowshade}{gray}{0.92}
\definecolor{headershade}{gray}{0.95}
\begin{document}

\title{Keyframe-Anchored Identity Preservation for Sequential-Action Video Generation
}

\author{Zhenjie Liu}
\affiliation{
  \institution{University of Science and Technology of China}
  \city{Hefei}
  \country{China}
  }
\email{liuzhenjie@mail.ustc.edu.cn}

\author{Binyan Chen}
\affiliation{
  \institution{University of Science and Technology of China}
  \city{Hefei}
  \country{China}
  }
\email{sztn@mail.ustc.edu.cn}

\author{Hao Chen}
\affiliation{
  \institution{University of Science and Technology of China}
  \city{Hefei}
  \country{China}
  }
\email{chhhenhao@mail.ustc.edu.cn}

\author{Tong Pan}
\affiliation{
  \institution{University of Science and Technology of China}
  \city{Hefei}
  \country{China}
  }
\email{pantong@mail.ustc.edu.cn}

\author{Shangfei Wang}
\authornote{The corresponding author.}
\affiliation{
  \institution{University of Science and Technology of China}
  \city{Hefei}
  \country{China}
  }
\email{sfwang@ustc.edu.cn}

\begin{abstract}
Identity-preserving text-to-video generation aims to synthesize a video that accurately follows a textual description while maintaining the recognizability of a user-specified subject throughout. The IPVG26 challenge extends this framework from a single holistic prompt to a temporally structured specification. The model additionally receives a sequence of timestamped action captions and must render the subject performing these actions in the specified order. This temporal structure presents a challenge not encountered in previous identity-preserving generation tasks, as the subject must continuously perform a scripted sequence of distinct actions while maintaining a consistent identity. However, end-to-end video generators are prone to appearance drift as motion accumulates and the depicted actions change. We address this challenge with a training-free, three-stage pipeline framework. An action-aware prompt polishment stage first rewrites the inputs into image-generation prompts that specify the terminal state of each action. An identity-preserving generation stage then produces the keyframe sequence by conditioning each frame jointly on the reference identity and its predecessor, thereby decoupling time-invariant appearance from time-varying pose. Finally, an identity-aware inference enhancement stage synthesizes the intermediate segments using multi-reference guidance and identity-driven noise searching, both of which reinforce identity fidelity during sampling. Our method ranked third on the official Track 2 leaderboard, demonstrating competitive performance and strong generality.
\end{abstract}



\begin{teaserfigure}
  \includegraphics[width=\textwidth]{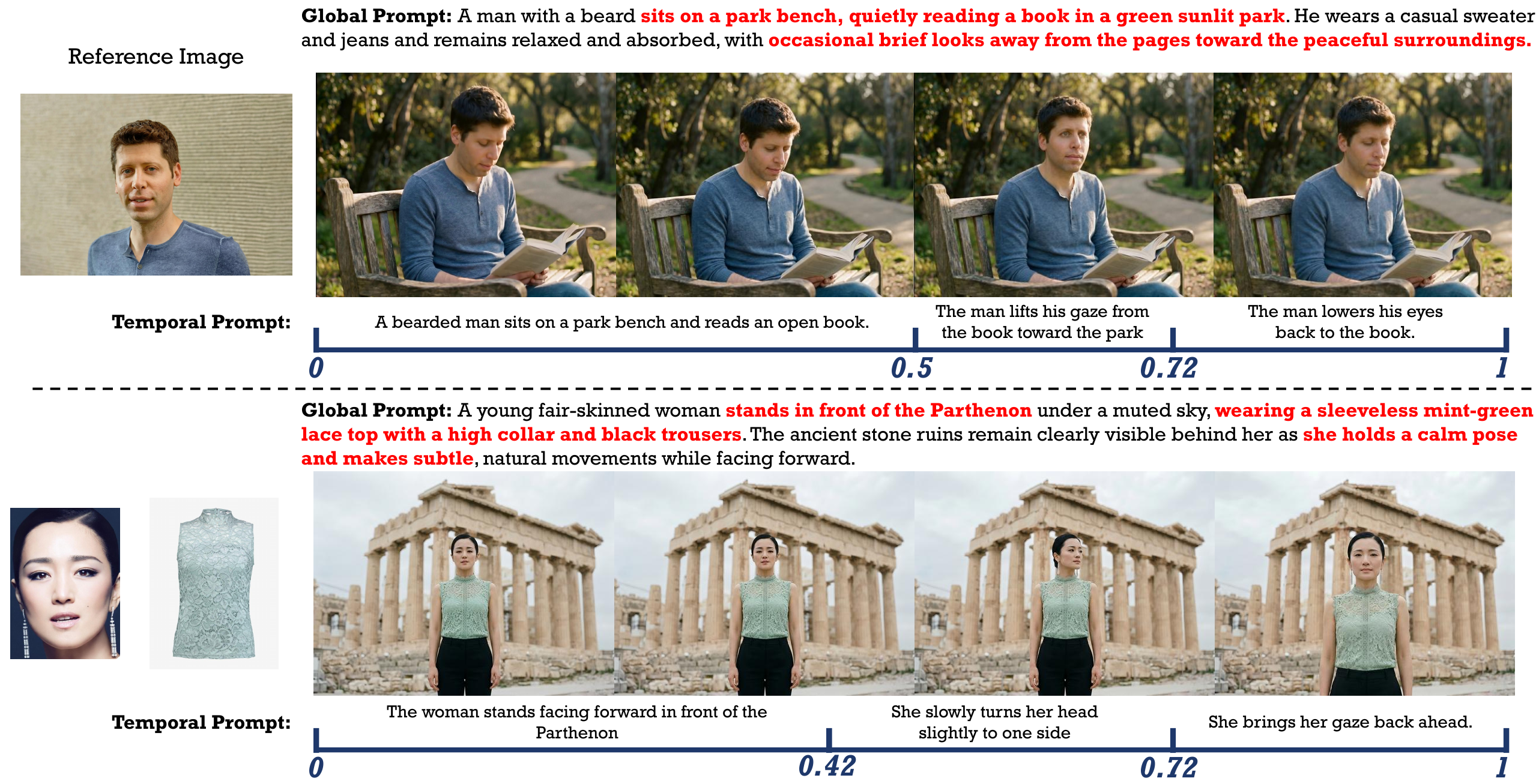}
  \caption{Examples of sequential action identity-preserving text-to-video generation by our framework. The track requires models to generate videos from reference images and structured timeline prompts that specify multiple sub-actions with precise timestamps. {\color{red}Red} in the text highlights key elements in instructions.}
  \label{fig:teaser}
\end{teaserfigure}

\maketitle

\section{Introduction}

Diffusion-based generative models have rapidly advanced text-to-video synthesis, and contemporary systems now generate open-domain scenes with convincing motion dynamics and high visual fidelity~\cite{ho2022video,blattmann2023stable,yang2024cogvideox,wan2025wan,Zheng2024OpenSoraDE}. As the quality of raw generation improves, the primary challenge increasingly shifts from whether a model can produce a plausible video to how precisely a user can control the content it depicts, ensuring the output reflects a specific intent rather than a generic prompt. Among the various dimensions of controllability, binding the generated video to a user-specified subject is one of the most critical. Identity-preserving text-to-video generation addresses this need: given a reference image of a subject and a textual description, it synthesizes a video that follows the description while maintaining the subject’s recognizability in every frame~\cite{yuan2025consisid,liu2025phantom,he2024idanimator}. This capability supports applications ranging from personalized content creation to controllable digital humans, where viewers must perceive the same individual throughout the clip rather than someone merely similar.

Track 2 of the IPVG26 challenge~\cite{panidentity} advances this setting from a single holistic prompt to a temporally structured specification. In addition to the reference image and a global scene description, the model receives a sequence of timestamped action captions and must render the subject performing these actions in the prescribed order. This temporal structure introduces a challenge that prior identity-preserving generation methods do not address, as the subject must move continuously through a scripted sequence of distinct actions while maintaining a consistent identity. End-to-end video generators are ill-suited for this task, as appearance tends to drift once motion accumulates and the depicted action changes, causing the subject to gradually lose its likeness over the course of a multi-action sequence.

We address this challenge with a keyframe-anchored decomposition. Instead of generating the entire video in a single pass, we represent the action sequence as a chain of identity-preserving keyframes, each anchored at the terminal state of an action segment. The video is then synthesized by interpolating between consecutive keyframes. This approach is implemented as a training-free, three-stage pipeline that requires no fine-tuning of off-the-shelf models: (1) an action-aware prompt refinement stage rewrites the inputs into image-generation prompts specifying the terminal state of each action; (2) an identity-preserving generation stage produces the keyframe chain by conditioning each frame jointly on the reference identity and its predecessor, thereby decoupling time-invariant appearance from time-varying pose; and (3) an identity-aware inference enhancement stage synthesizes the intervening segments using multi-reference guidance and identity-driven noise searching~\cite{hyung2024stg,liu2025consisttalk}, both of which reinforce identity fidelity during sampling.

The proposed pipeline ranked third on the official Track 2 leaderboard, with its most significant gains concentrated on the heavily weighted identity-preservation and temporal-alignment metrics. This outcome corroborates the effectiveness of the keyframe-anchored design. The main contributions of this work are summarized as follows:

\begin{itemize}
    \item We propose an \textbf{Action-Aware Prompt Polishment} module that rewrites the global scene description and temporal action captions into terminal-state image-generation prompts, providing clear and unambiguous spatial targets for keyframe synthesis.
    \item We design an \textbf{ID-Preserving Chained Keyframe Generation} stage that conditions each keyframe jointly on the reference identity and the preceding keyframe, effectively decoupling time-invariant identity from time-varying pose, thereby preserving both recognizability and motion continuity.
    \item We develop an \textbf{ID-Aware Inference Enhancement} stage that integrates multi-reference guidance, identity-driven noise searching, and a variable-length segment allocation strategy to reinforce identity fidelity during sampling, all without requiring additional training.
\end{itemize}

\begin{figure*}[t]
  \centering
  \includegraphics[width=\textwidth]{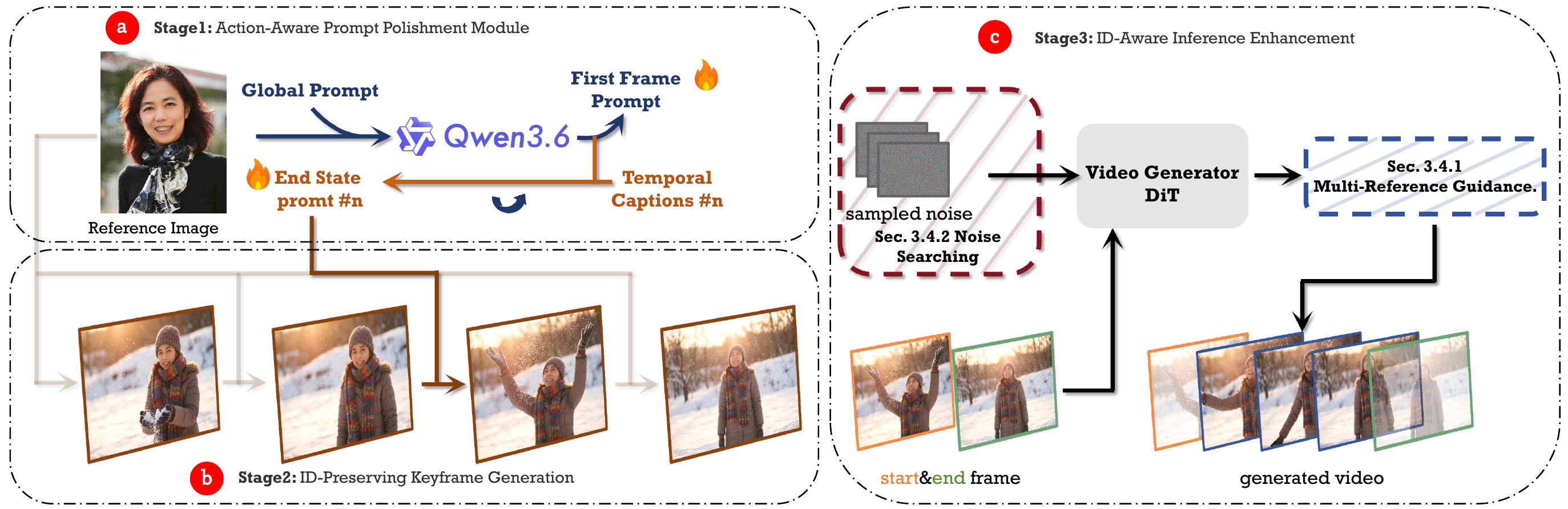}
  \caption{The proposed framework consists of three main modules. \textbf{(a)~Action-Aware Prompt Polishment} first combines the global scene description with the reference identity image into a first-frame prompt, and then rewrites each temporal caption into an end-frame prompt that defines the terminal state of each action segment. \textbf{(b)~ID-Preserving Keyframe Generation} conditions on $I_{\text{ref}}$ together with the polished prompts to generate a chained sequence of keyframes, maintaining identity and pose continuity across segments. \textbf{(c)~ID-Aware Inference Enhancement} synthesizes each segment from its boundary keyframes and temporal caption using Multi-Reference Guidance (\cref{sec:method:mrg}) and Noise Searching (\cref{sec:method:ns}) schemes, and finally concatenates the segments to produce the final video.}
  \label{fig:overview}
\end{figure*}

\section{Related Work}

\subsection{Identity-Preserving Text-to-Video Generation}
Early identity-preserving text-to-video methods bind a subject by fine-tuning model weights or inserting per-identity modules~\cite{hu2022lora,ma2025magic,wei2024dreamvideo}, which must be retrained for every new identity and therefore scale poorly. To eliminate this per-identity cost, tuning-free approaches attach a dedicated identity branch trained offline on paired data~\cite{ye2023ip,li2024photomaker,wang2024instantid}. Recent reference-conditioned video generators further enable open-domain, zero-shot subject generation~\cite{yuan2025consisid,liu2025phantom,zhong2025concat,hu2025hunyuancustom,fei2025skyreels,Wang2025IdentityPreservingTG,Zhang2025FantasyIDFK}. However, these models rely on large training corpora and costly post-training. Their end-to-end joint optimization of appearance and motion induces a spatiotemporal trade-off, causing identity drift as motion accumulates. Moreover, they are designed for a single holistic prompt rather than a scripted sequence of timed actions. Instead, we adopt a keyframe-decoupled paradigm that separates the time-invariant identity layout from temporal motion, which is cast as interpolation between consecutive keyframes. This separation sidesteps the trade-off and directly accommodates the sequential-action setting.

\subsection{Inference-time Enhancement}
Classifier-free guidance \cite{cfree-guidance} directs sampling toward the text condition but does not provide explicit control over subject identity. This limitation has motivated the development of identity-aware and spatiotemporal guidance variants that augment the score function with additional conditioning terms~\cite{Gao2025IdentityPreservingTG,hyung2024stg}. Building on this approach, our multi-reference guidance introduces a dual-constraint variant that couples the boundary keyframes with the text prompt, explicitly amplifying the identity-bearing visual condition. A complementary approach treats the initial diffusion noise as a searchable variable rather than a fixed draw, selecting an initialization through short lookahead denoising guided by a reward~\cite{liu2025consisttalk}. We adapt this concept for identity preservation by performing an identity-driven noise search across segments. Both enhancements are training-free and operate solely at sampling time, consistent with the rest of our pipeline.

\section{Method}

\subsection{Overview and Task Formulation}
\label{sec:method:overview}

Given a reference identity image $I_{\text{ref}}$, a global scene description $P_{\text{global}}$, and a temporal action sequence $\{(s_i, e_i, c_i)\}_{i=1}^{N}$, where $s_i, e_i \in [0,1]$ are normalized timestamps and $c_i$ is the action caption for the $i$-th segment, the goal of Track 2~\cite{panidentity} is to synthesize a video $V$ that simultaneously satisfies three requirements: the subject in $V$ must preserve the identity depicted in $I_{\text{ref}}$; the overall scene semantics must align with $P_{\text{global}}$ while each temporal segment faithfully reflects its corresponding caption $c_i$; and the transitions between consecutive segments must remain temporally coherent.

We address this task using a three-stage pipeline, illustrated in \cref{fig:overview}. First, the \emph{Action-Aware Prompt Polishment} module rewrites $P_{\text{global}}$ and each caption $c_i$ into descriptions tailored for image generation, emphasizing the terminal state of every action segment (\cref{sec:method:prompt}). Second, the \emph{ID-Preserving Keyframe Generation} stage produces a chained sequence of keyframes $\{K_0, K_1, \ldots, K_N\}$ by conditioning each frame on $I_{\text{ref}}$ along with its predecessor, thereby maintaining both identity and motion continuity (\cref{sec:method:keyframe}). Finally, the \emph{ID-Aware Inference Enhancement} stage synthesizes the output video from these keyframes, applying multi-reference guidance and identity-driven noise searching to reinforce identity preservation, while a variable-length strategy allocates the frame budget across segments proportionally to their durations (\cref{sec:method:inference}).

\subsection{Action-Aware Prompt Polishment}
\label{sec:method:prompt}

The textual inputs for Track 2, specifically a global scene description $P_{\text{global}}$ and a set of per-segment action captions $\{c_i\}_{i=1}^{N}$, are written as video-level narratives and therefore lack the spatial precision and identity specificity required by image generation models. To address this issue, we use a multimodal large language model (LLM) to perform a controlled rewrite of each caption into an image-generation prompt, conditioned on the reference image $I_{\text{ref}}$. As illustrated in panel (a) of \cref{fig:overview}, this refinement process occurs in two stages. In the first stage, $P_{\text{global}}$ is rewritten into a first-frame prompt $\hat{P}_{\text{global}}$ that anchors the subject's identity attributes (face shape, skin tone, hairstyle, and distinctive facial traits) as observed in $I_{\text{ref}}$, while establishing the scene composition, camera framing, and lighting atmosphere. The rewritten prompt is required to depict a plausible opening moment that implies the onset of motion through pose and spatial layout, rather than collapsing into a generic portrait. In the second stage, each temporal caption $c_i$ is independently transformed into an end-frame prompt $\hat{c}_i$ that specifies the \emph{terminal state} of the $i$-th action segment. The instruction explicitly directs the LLM to describe the body pose, hand position, and gaze direction at the moment the segment concludes, while preserving spatial continuity with the preceding frame.

The emphasis on terminal states serves a clear functional purpose. Since our downstream keyframe generator (\cref{sec:method:keyframe}) produces the ending frame of each segment, the polished prompt must specify the final position of the action rather than its progression over time. This approach provides unambiguous spatial constraints that facilitate chained keyframe generation and promote smooth transitions between segments. The instruction template for temporal caption polishing is summarized below.

\begin{quote}\small
\textbf{Inputs}\\
1.~Reference identity image $I_{\text{ref}}$\\
2.~Temporal caption $c_i$ (action description with normalized start/end timestamps)\\[2pt]
\textbf{Task}\\
Rewrite $c_i$ into a single image-generation prompt for the end frame of this segment that
(i)~preserves the subject's identity from $I_{\text{ref}}$, including face shape, skin tone, age, appearance, and hairstyle;
(ii)~depicts the resulting state of the action, expressing body pose, hand position, and gaze direction after the segment;
(iii)~maintains spatial continuity with the body orientation of the previous frame;
(iv)~reads as a natural, descriptive sentence suitable for image generation.\\[2pt]
\textbf{Output}\\
A single polished end-frame prompt $\hat{c}_i$.
\end{quote}

\subsection{ID-Preserving Keyframe Generation}
\label{sec:method:keyframe}

Given the polished prompts, we generate a sequence of keyframes $\{K_0, K_1, \ldots, K_N\}$ with an image generation model in a strictly sequential, chained manner, as illustrated in panel (b) of \cref{fig:overview}. We produce the first keyframe $K_0$ by conditioning the model on the reference image $I_{\text{ref}}$ and the polished global prompt $\hat{P}_{\text{global}}$, yielding a scene that depicts the subject in a neutral initial pose before any action unfolds. Each subsequent keyframe $K_i$ ($i \geq 1$) is generated from three inputs: the reference image $I_{\text{ref}}$, a persistent identity anchor; the preceding keyframe $K_{i-1}$, a visual continuity cue; and the polished end-frame prompt $\hat{c}_i$, which specifies the target terminal state of the $i$-th action segment. We instruct the model to inherit the pose, position, and orientation from $K_{i-1}$ and modify only the aspects explicitly described by $\hat{c}_i$, producing a natural progression of motion across the keyframe sequence.

This chained conditioning design separates two orthogonal constraints by assigning each to a distinct input. The reference image $I_{\text{ref}}$ supplies time-invariant identity features, including facial structure, skin tone, and hairstyle, ensuring the subject remains recognizable throughout the sequence despite pose changes. In contrast, the preceding keyframe $K_{i-1}$ encodes the time-varying spatial state, encompassing body orientation, limb configuration, and scene layout at the end of the previous segment. By supplying both signals simultaneously, the model maintains identity without compromising temporal coherence, and each transition reflects only the incremental change dictated by the corresponding action caption.

\subsection{Video Generation and ID-Aware Inference Enhancement}
\label{sec:method:inference}

With the keyframe sequence $\{K_0, K_1, \ldots, K_N\}$ established, the video generation stage, shown in panel~(c) of \cref{fig:overview}, synthesizes each action segment by conditioning a diffusion-based video model on the corresponding pair of boundary keyframes, which serve as the first-frame and last-frame constraints. Because the temporal captions define segments of unequal duration, we adopt a variable-length allocation strategy: the shortest segment is assigned a base frame budget $F_{\min}$, and every other segment receives a frame count proportional to its normalized duration, rounded to the nearest value that satisfies the temporal VAE constraint of $8k+1$. The resulting segment videos are concatenated at their shared boundary frames, with the first frame of each non-initial segment discarded to avoid duplication. Beyond this generation pipeline, we introduce two inference-time training-free enhancements that explicitly optimize identity preservation.

\subsubsection{Multi-Reference Guidance}
\label{sec:method:mrg}

Standard classifier-free guidance (CFG)~\cite{cfree-guidance} directs the denoising process toward text-conditioned predictions but does not provide an explicit mechanism to enforce identity consistency with the reference subject. Inspired by recent work on identity-aware guidance~\cite{Gao2025IdentityPreservingTG}, we extend the sampling procedure with a multi-modal guidance formulation that simultaneously leverages three complementary signals. Let $\epsilon_\theta(x_t \mid \mathcal{C})$ denote the noise prediction conditioned on the full context $\mathcal{C}$ that comprises the text prompt, the boundary keyframes, and the reference identity. The guided prediction at each denoising step takes the form
\begin{equation}
\label{eq:guidance}
\tilde{\epsilon}_\theta(x_t) = \epsilon_\theta(x_t \mid \mathcal{C}) + w_{\text{cfg}}\,\Delta_{\text{cfg}} + w_{\text{stg}}\,\Delta_{\text{stg}} + w_{\text{mod}}\,\Delta_{\text{mod}}
\end{equation}
where $\Delta_{\text{cfg}} = \epsilon_\theta(x_t \mid \mathcal{C}) - \epsilon_\theta(x_t \mid \varnothing_{\text{text}})$ is the standard CFG term that enforces semantic alignment with the text prompt, $\Delta_{\text{stg}} = \epsilon_\theta(x_t \mid \mathcal{C}) - \epsilon_\theta^{\text{perturb}}(x_t \mid \mathcal{C})$ is a spatiotemporal guidance term obtained by perturbing selected attention layers to encourage temporal coherence, and $\Delta_{\text{mod}} = \epsilon_\theta(x_t \mid \mathcal{C}) - \epsilon_\theta(x_t \mid \varnothing_{\text{img}})$ is a modality guidance term that amplifies the influence of the visual conditioning, namely the boundary keyframes that carry the identity information. The three scales $w_{\text{cfg}}$, $w_{\text{stg}}$, and $w_{\text{mod}}$ control the relative strength of each signal. By explicitly separating the identity-bearing visual condition from the text condition, this formulation guides the sampling trajectory toward identity-faithful outputs.

\begin{table*}[!ht]
\begin{center}
    \caption{Quantitative comparison of the two video generation backbones we explored on the IPVG26 Track~2 test set ($200$ samples). All scores lie in $[0,1]$ and higher is better ($\uparrow$). The best result in each column is marked in \textbf{bold}.}
    \label{tab:leaderboard}
    \vspace{-2mm}
\resizebox{0.99\linewidth}{!}{
\begin{tabular}{l|cccc|cc|cc|cc|c}
\toprule[1.5pt]
\multirow{2}{*}{\textbf{Backbone}} & \multicolumn{4}{c|}{Identity} & \multicolumn{2}{c|}{Global Alignment} & \multicolumn{2}{c|}{Temporal Alignment} & \multicolumn{2}{c|}{Quality} & \multicolumn{1}{c}{Overall} \\
\cline{2-12}
\specialrule{0em}{1pt}{1pt}
& \textbf{Cur}$\uparrow$ & \textbf{Arc}$\uparrow$ & \textbf{DINO}$\uparrow$ & \textbf{ID}$\uparrow$ & \textbf{CLIP}$\uparrow$ & \textbf{GME}$\uparrow$ & \textbf{CLIP}$\uparrow$ & \textbf{GME}$\uparrow$ & \textbf{MS}$\uparrow$ & \textbf{IQ}$\uparrow$ & \textbf{O.S.}$\uparrow$ \\
\midrule
Wan2.1-VACE & 0.4503 & 0.4187 & 0.1289 & 0.3861 & 0.2918 & 0.6104 & 0.2549 & 0.5093 & 0.9847 & 0.5921 & 0.4818 \\
\rowcolor{rowshade}
LTX-2.3 & \textbf{0.4731} & \textbf{0.4392} & \textbf{0.1402} & \textbf{0.4055} & \textbf{0.3046} & \textbf{0.6293} & \textbf{0.2654} & \textbf{0.5266} & \textbf{0.9883} & \textbf{0.6078} & \textbf{0.4971} \\
\bottomrule[1.5pt]
\end{tabular}
}
\vspace{-2pt}
\end{center}
\end{table*}

\subsubsection{Noise Searching}
\label{sec:method:ns}

The initial noise $z_T$ drawn at the start of the reverse diffusion process significantly influences the identity fidelity of the generated video. Inspired by IC-Init~\cite{liu2025consisttalk}, we introduce an identity-driven noise search procedure that treats the initial noise as a variable to be optimized rather than a fixed random draw. For each video segment, we sample $K$ candidate noise tensors $\{z_T^{(1)}, \ldots, z_T^{(K)}\}$ and perform a short lookahead denoising of $T'$ steps ($T' \ll T$) on each candidate to obtain partially denoised latents. We then decode the first frame of each candidate and compute a reward based on the cosine similarity between its face embedding and that of the reference image $I_{\text{ref}}$:
\begin{equation}
\label{eq:noise_reward}
R(z_T^{(k)}) = \mathrm{sim}\!\left(\phi(D(z_{T-T'}^{(k)})),\; \phi(I_{\text{ref}})\right)
\end{equation}
where $D(\cdot)$ denotes the VAE decoder and $\phi(\cdot)$ extracts a face embedding. The candidate that attains the highest reward is retained for full denoising. This procedure converts the stochastic noise initialization into a constrained search that favors identity-preserving diffusion trajectories, at the cost of $K$ additional short forward passes per segment.

\begin{table}[t]
\begin{center}
    \caption{Final standings on the official IPVG26 Track 2 leaderboard. The reported score is the official ranking metric. Our entry (USTC-AC) is highlighted.}
    \label{tab:ranking}
    \vspace{-2mm}
\begin{tabular}{c|l|c}
\toprule[1.5pt]
\textbf{Rank} & \textbf{Team} & \textbf{Score}$\downarrow$ \\
\midrule
1 & USTC-CMI & 1.25 \\
2 & WislabGDUT & 1.8125 \\
\rowcolor{rowshade}
3 & USTC-AC (Ours) & 2.9375 \\
\bottomrule[1.5pt]
\end{tabular}
\vspace{-2pt}
\end{center}
\end{table}

\section{Experiments}

\subsection{Implementation Details}
\label{sec:exp:impl}

\paragraph{Models and configuration.} Our pipeline integrates three components. The \textbf{action-aware prompt polishment} stage is implemented with Qwen3.6-27B~\cite{yang2025qwen3}, a multimodal large language model that, conditioned on the reference image, rewrites the global prompt and the temporal captions. The \textbf{ID-preserving keyframes} are synthesized by a locally deployed Z-Image~\cite{zimage2025} model at a resolution of $1536 \times 1024$. For \textbf{video synthesis}, we adopt LTX-2.3~\cite{hacohen2024ltxvideo,HaCohen2026LTX2EJ} in its keyframe-interpolation mode, conditioning each segment on the corresponding pair of boundary keyframes. Additionally, we consider Wan2.1-VACE~\cite{wan2025wan,jiang2025vace} for first- and last-frame guided generation. Videos are produced at $24$ frames per second with $40$ denoising steps. Under the variable-length allocation strategy, the shortest segment is assigned $F_{\min} = 121$ frames, while the remaining segments are scaled proportionally and rounded to the nearest valid length of the form $8k+1$. The multi-reference guidance scales are set to $w_{\text{cfg}} = 5.0$, $w_{\text{stg}} = 1.5$ (perturbing transformer block $29$), and $w_{\text{mod}} = 3.0$.

\paragraph{Evaluation protocol.} We evaluate our method on the official Track 2 test set~\cite{panidentity}, which is built upon the ReactID dataset~\cite{li2026reactid} and consists of 200 samples. Each sample includes one or more reference images, a category label, a global scene prompt, and a sequence of temporally segmented action captions. Generation quality is assessed across three complementary dimensions. \textbf{Identity preservation} is measured using an identity score tailored to the subject category: facial subjects are evaluated with CurricularFace~\cite{huang2020curricularface} and ArcFace~\cite{deng2019arcface} embeddings, while object and animal subjects are assessed using DINOv2~\cite{oquab2023dinov2} features. \textbf{Text-video alignment} is quantified by CLIP~\cite{radford2021learning} and GME~\cite{zhang2024gme} similarities computed at both the global and per-segment temporal level; $32$ frames are uniformly sampled for the global measurement, and $8$ frames are sampled within each segment. \textbf{Visual fidelity} is captured by motion smoothness (MS) and image quality (IQ)~\cite{huang2024vbench}. The overall score aggregates these metrics through a weighted sum,
\begin{equation}
\label{eq:overall}
\begin{aligned}
S_{\text{overall}} = {}& 0.20\,S_{\text{id}} + 0.15\,S_{\text{gme}}^{\text{g}} + 0.10\,S_{\text{clip}}^{\text{g}} \\
& + 0.25\,S_{\text{gme}}^{\text{t}} + 0.15\,S_{\text{clip}}^{\text{t}} + 0.075\,S_{\text{ms}} + 0.075\,S_{\text{iq}},
\end{aligned}
\end{equation}
where the superscripts $\text{g}$ and $\text{t}$ denote the global and temporal variants of the alignment metrics, respectively.

\begin{table}[!t]
\begin{center}
    \caption{Ablation study of the two inference-time enhancements, Multi-Reference Guidance (MRG) and Noise Searching (NS), on the Track 2 test set. Each variant removes one component from the full LTX-2.3 model (highlighted). The best score per column is in \textbf{bold}.}
    \label{tab:ablation}
    \vspace{-2mm}
\resizebox{\linewidth}{!}{
\begin{tabular}{l|c|cc|cc|c}
\toprule[1.5pt]
\multirow{2}{*}{\textbf{Variant}} & \multicolumn{1}{c|}{Identity} & \multicolumn{2}{c|}{Alignment} & \multicolumn{2}{c|}{Quality} & \multicolumn{1}{c}{Overall} \\
\cline{2-7}
\specialrule{0em}{1pt}{1pt}
& \textbf{ID}$\uparrow$ & \textbf{G-GME}$\uparrow$ & \textbf{T-GME}$\uparrow$ & \textbf{MS}$\uparrow$ & \textbf{IQ}$\uparrow$ & \textbf{O.S.}$\uparrow$ \\
\midrule
\rowcolor{rowshade}
Full model & \textbf{0.4055} & \textbf{0.6293} & \textbf{0.5266} & \textbf{0.9883} & \textbf{0.6078} & \textbf{0.4971} \\
\quad w/o MRG & 0.3520 & 0.6248 & 0.5176 & 0.9870 & 0.6011 & 0.4820 \\
\quad w/o NS & 0.3853 & 0.6270 & 0.5223 & 0.9879 & 0.6052 & 0.4910 \\
\bottomrule[1.5pt]
\end{tabular}
}
\vspace{-2pt}
\end{center}
\end{table}

\subsection{Quantitative and Qualitative Analysis}
\label{sec:exp:results}

\begin{figure*}[t]
  \centering
  \includegraphics[width=\textwidth]{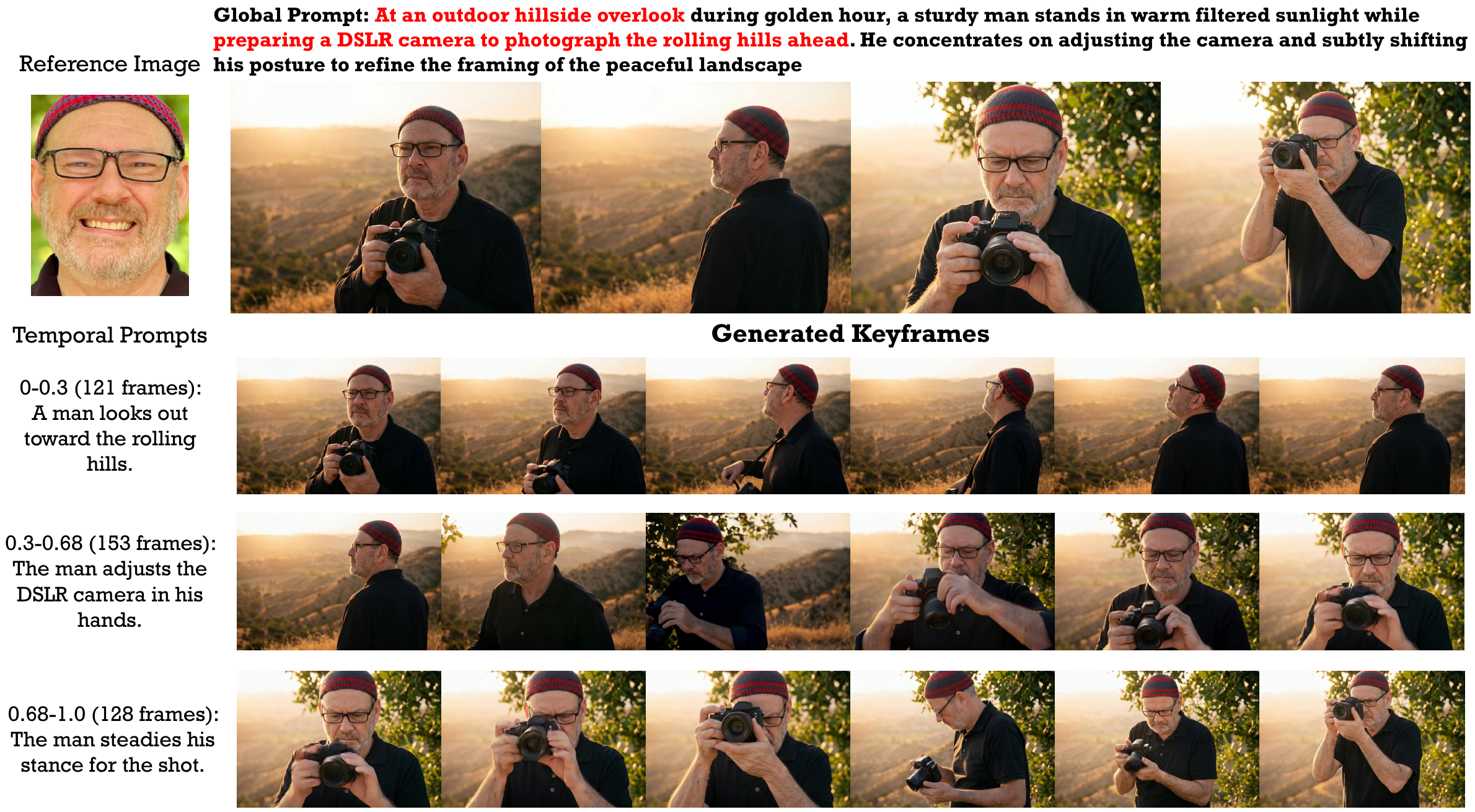}
  \caption{\textbf{Qualitative case study.} From the reference identity and the global prompt, our pipeline generates the chained keyframes $K_0$--$K_3$ and synthesizes the three action segments, each shown with its temporal caption and a uniformly sampled frame sequence.}
  \label{fig:qualitative}
\end{figure*}

\paragraph{Quantitative results.}
During development, we experimented with two candidate video generation backbones within an identical keyframe pipeline: Wan2.1-VACE in first- and last-frame mode, and LTX-2.3 in keyframe-interpolation mode. \cref{tab:leaderboard} presents their performance on the Track 2 test set. LTX-2.3 outperforms Wan2.1-VACE across all seven evaluated metrics, increasing the overall score from $0.4818$ to $0.4971$, which motivated us to adopt it as our final backbone. Motion smoothness is already near saturation for both backbones (above $0.98$), so the discriminative signal primarily resides in the identity and alignment metrics, where LTX-2.3 consistently holds an advantage. \cref{tab:ranking} reports the final standings on the official Track 2 leaderboard, where our entry ranked third, confirming the effectiveness of the proposed pipeline.

\paragraph{Qualitative results.}
\cref{fig:qualitative} presents a representative case study. Across the four chained keyframes and all sampled frames, the man's facial structure, skin tone, and hairstyle remain consistent with the reference image. Each segment converges on the terminal state described by its temporal caption, progressing from looking out toward the hills to adjusting the DSLR in hand, and finally steadying the stance for the shot. This behavior validates our end-frame prompt design, which anchors every keyframe at the point where the action lands rather than how it unfolds over time. Because each keyframe inherits the pose and orientation of its predecessor, the boundaries between consecutive segments are visually continuous and free of abrupt jumps. The synthesized content also adheres to the global prompt, faithfully reproducing the golden-hour hillside overlook and the photographic activity it describes.

\subsection{Ablation Study}
\label{sec:exp:ablation}

\noindent We ablate the two inference-time enhancements introduced in \cref{sec:method:inference} by removing each component individually from the full LTX-2.3 model and report the results in \cref{tab:ablation}. Removing Multi-Reference Guidance causes the most significant degradation, with the identity score dropping from $0.4055$ to $0.3520$. This confirms that explicitly amplifying the identity-bearing visual condition is the primary driver of subject consistency, as text and keyframe conditioning alone do not sufficiently steer the sampling trajectory toward the reference appearance. Removing Noise Searching has a milder but consistent effect, lowering the identity score to $0.3853$ and the overall score from $0.4971$ to $0.4910$. Serving as a refinement that selects an identity-favorable initialization on top of the guided sampler, it contributes a smaller yet reliable gain. Both components leave motion smoothness and image quality nearly unchanged, indicating that their benefits are concentrated on identity preservation. Taken together, these two enhancements are complementary and jointly account for the improvements in the most heavily weighted evaluation dimensions.

\section{Conclusion}

We propose a keyframe-anchored approach for identity-preserving video generation in the IPVG26 Track 2 setting, where a subject performs a scripted sequence of timestamped actions while remaining recognizable. Instead of synthesizing the entire clip at once, we decompose the action sequence into identity-preserving keyframes anchored at segment endpoints and interpolate between them. Our training-free, three-stage pipeline requires no fine-tuning of off-the-shelf models: (1) an action-aware prompt polishing stage that rewrites global and temporal captions into terminal-state image prompts; (2) a chained keyframe generation stage that conditions each frame on the reference identity and its predecessor, effectively separating appearance from pose; and (3) an identity-aware inference enhancement stage that synthesizes intermediate frames using multi-reference guidance and identity-driven noise searching. Our entry ranked third on the official leaderboard, excelling in identity preservation and temporal alignment. Ablation studies show that inference-time enhancements, especially multi-reference guidance, drive subject consistency. These results demonstrate that keyframe-anchored decomposition is an effective method for identity-preserving multi-action video generation.

\clearpage
\begin{acks}
This work was supported by the National Natural Science Foundation of China 62376255.
\end{acks}

\bibliographystyle{ACM-Reference-Format}
\bibliography{main}

\end{document}